\documentclass[runningheads]{llncs}
\usepackage[T1]{fontenc}
\usepackage{graphicx,verbatim}
\usepackage{hyperref}

\usepackage{amsmath}
\usepackage{amssymb}
\usepackage{booktabs}
\usepackage{xspace}
\usepackage{makecell}
\usepackage{caption}
\usepackage{subcaption}
\usepackage{xcolor}

\newcommand{\ourmodel}{BLINK\xspace}

\begin{document}
%
\title{BLINK: Behavioral Latent Modeling \\of NK Cell Cytotoxicity}
%
\author{Iman Nematollahi\inst{1} \and
Jose Francisco Villena-Ossa\inst{2} \and
Alina Moter\inst{3} \and \\
Kiana Farhadyar\inst{1,5} \and
Gabriel Kalweit\inst{4,1} \and
Abhinav Valada\inst{1} \and
Toni Cathomen\inst{2} \and
Evelyn Ullrich\inst{3} \and
Maria Kalweit\inst{1,4,5}
}
\authorrunning{I. Nematollahi et al.}
%
\institute{Department of Computer Science, University of Freiburg, Freiburg, Germany \and
Institute for Transfusion Medicine and Gene Therapy, University Medical Center Freiburg, Freiburg, Germany \and
Goethe University, Department of Pediatrics, Experimental Immunology and Cell Therapy, Frankfurt am Main, Germany \and
Collaborative Research Institute Intelligent Oncology (CRIION), Freiburg, Germany \and
IMBIT//BrainLinks-BrainTools, University of Freiburg, Freiburg, Germany}

  
\maketitle              
\begin{abstract}
Machine learning models of cellular interaction dynamics hold promise for understanding cell behavior. Natural killer (NK) cell cytotoxicity is a prominent example of such interaction dynamics and is commonly studied using time-resolved multi-channel fluorescence microscopy.
Although tumor cell death events can be annotated at single frames, NK cytotoxic outcome emerges over time from cellular interactions and cannot be reliably inferred from frame-wise classification alone. We introduce BLINK, a trajectory-based recurrent state-space model that serves as a cell world model for NK–tumor interactions. BLINK learns latent interaction dynamics from partially observed NK–tumor interaction sequences and predicts apoptosis increments that accumulate into cytotoxic outcomes. Experiments on long-term time-lapse NK–tumor recordings show improved cytotoxic outcome detection and enable forecasting of future outcomes, together with an interpretable latent representation that organizes NK trajectories into coherent behavioral modes and temporally structured interaction phases. BLINK provides a unified framework for quantitative evaluation and structured modeling of NK cytotoxic behavior at the single-cell level.

\end{abstract}
\section{Introduction}
Natural killer (NK) cells are cytotoxic lymphocytes of the innate immune system that play a central role in tumor immunosurveillance and emerging cellular immunotherapies, including chimeric antigen receptor (CAR)–engineered NK cells~\cite{yokoyamaDynamicLifeNatural2004,wendelArmingImmuneCells2021,bexteEngineeringPotentCAR2024}. Their cytotoxic activity arises from dynamic, context-dependent interactions with tumor cells, involving migration, target engagement, contact formation, and apoptosis induction, which is a regulated form of programmed cell death \cite{lanierNKCellRecognition2005,ramirez-labradaAllNKCellMediated2022,moterMigrationDynamicsHuman2025a}. Accurate assessment of NK efficacy is essential for evaluating immune competence and optimizing engineered products~\cite{sordo-bahamondeMechanismsResistanceNK2020}. Since cytotoxic outcomes arise from dynamic interaction processes rather than instantaneous binary events, distinguishing effective from ineffective NK–tumor interactions requires high-resolution, time-resolved single-cell analysis~\cite{alievaBEHAV3D3DLive2024}. However, conventional assays rely on bulk or terminal measurements, or on expert visual inspection and manual annotation of trajectories, limiting scalability and obscuring the temporal structure and heterogeneity of individual NK interactions~\cite{zhuCytotoxicChemotacticDynamics2023}. A trajectory-level framework for quantifying NK-induced tumor cell death is therefore critical for linking dynamic interaction behavior to cytotoxic outcome.

Time-resolved fluorescence microscopy enables direct observation of NK--tumor co-cultures, providing multi-channel measurements of morphology, cell identity, and apoptotic signals~\cite{deguineIntravitalImagingReveals2010,vanherberghenClassificationHumanNatural2013}. While tumor cell death events can be annotated at the frame level, modeling cytotoxic outcome as time-independent frame-wise classifications neglects the structured interaction dynamics underlying NK-induced apoptosis. Cytotoxic outcome is inherently monotonic and evolves over time~\cite{cerignoliVitroImmunotherapyPotency2018}, driven by latent states reflecting contact history and intracellular processes. Effective evaluation therefore requires models that capture latent interaction dynamics and produce coherent estimates of cumulative cytotoxic outcome.\looseness=-1

This perspective aligns with the emerging vision of a virtual cell~\cite{bunneHowBuildVirtual2024}: a computational model that infers cellular state and predicts its evolution from observational data, reducing reliance on costly experimental inspection and manual trajectory assessment. World models~\cite{haWorldModels2018} provide a principled framework for this paradigm by learning latent dynamical representations from sequential observations. By encoding observations into a compact state and modeling its temporal dynamics, world models enable inference and forecasting in partially observable systems. Widely used in reinforcement learning~\cite{hafnerMasteringAtariDiscrete2020} and robotics~\cite{nematollahiLUMOSLanguageConditionedImitation2025,chandraDiWADiffusionPolicy2025} to model environment dynamics from image sequences, world models provide a natural framework for NK cytotoxic outcome modeling, where apoptosis is not directly observable but emerges from interaction histories between NK and tumor cells. Inferring this latent cellular condition from time-resolved morphological observations enables structured prediction of cytotoxic outcome trajectories.

In this work, we propose a behaviorally grounded latent dynamical framework that infers latent interaction states from NK behavior and uses them to model cytotoxic outcome over time. We instantiate this perspective in \textbf{\ourmodel}, a trajectory-based recurrent state-space model serving as a cell world model for estimating cumulative cytotoxic outcome from time-resolved microscopy. The architecture builds on a DreamerV2-inspired latent state-space model~\cite{hafnerMasteringAtariDiscrete2020} to capture interaction dynamics and augments it with a biologically grounded prediction head that estimates cytotoxic outcome increments. We make three contributions:
(i) we formalize cumulative NK cytotoxic outcome estimation as inference over latent interaction dynamics rather than frame-wise time-independent event classification; (ii) we introduce an action-conditioned recurrent state-space world model that captures structured interaction dynamics from morphology, motion, and apoptotic signals; and (iii) we demonstrate that this formulation improves cumulative cytotoxic outcome prediction, enables forecasting of future outcome, and yields an interpretable latent representation that organizes NK trajectories into coherent behavioral modes and temporally structured interaction phases. To the best of our knowledge, \ourmodel is the first to employ a latent recurrent state-space world model to time-lapse fluorescence microscopy, establishing a unified framework for structured modeling of single-cell interaction dynamics and functional outcomes.\looseness=-1
\section{Problem Formulation}
We investigate the problem of estimating cumulative NK-induced tumor cell death from time-resolved fluorescence microscopy. We assume access to multi-channel time-resolved microscopy recordings of NK–tumor co-cultures, comprising brightfield morphology, NK and tumor fluorescence, and a viability channel. Formally, a recording is represented as $X = (X_0, \dots, X_T)$, where $X_t \in \mathbb{R}^{H \times W \times C}$ denotes the multi-channel image at time $t$. From these recordings, segmentation and tracking are employed to extract NK–tumor interaction trajectories. For each tracked NK cell, we generate image crops centered on the NK cell at each time step, yielding a dataset $\mathcal{D} = \{\tau^{(i)}\}_{i=1}^{N}$, where each trajectory $\tau^{(i)} = (x^{(i)}_0, \dots, x^{(i)}_{T_i})$ corresponds to a temporally ordered sequence of interaction crops of length $T_i$. Frame-level tumor cell death annotations are derived from a caspase-activated viability channel, labeling each tumor cell as apoptotic at the first frame where its signal exceeds a predefined threshold.\looseness=-1

We model NK--tumor interactions as a partially observable Markov decision process $\mathcal{M} = (\mathcal{S}, \mathcal{A}, \mathcal{X}, \mathcal{P})$, where $s_t \in \mathcal{S}$ denotes the latent interaction state, $a_t \in \mathcal{A}$ represents the NK cell 2D displacement in the imaging plane between frames, $x_t \in \mathcal{X}$ denotes the observed multi-channel microscopy image, and $\mathcal{P}(s_{t+1} \mid s_t, a_t)$ governs the latent interaction dynamics. The interaction state is not directly observable, and cytotoxic outcomes arise as consequences of these latent dynamics. This formulation motivates learning a latent state model that infers and propagates interaction dynamics from partial observations and actions to support temporally consistent prediction of cytotoxic outcomes.

Cytotoxic outcome is a monotonic cumulative process evolving over time. We therefore formulate the task as estimating cumulative NK-induced tumor cell death over finite temporal windows. Let $y_t$ denote the cumulative NK-induced tumor cell death up to time $t$. For a window starting at time $t_0$ with length $L$, we define relative cumulative tumor cell death $\tilde{y}_t = y_t - y_{t_0}$ for $t \in \{t_0, \dots, t_0 + L-1\}$, where $\tilde{y}_{t_0} = 0$ and $\tilde{y}_{t+1} \ge \tilde{y}_t$. The objective is to estimate the cumulative progression $\tilde{y}_{t_0:t_0+L-1}$ from the observation history $x_{t_0:t_0+L-1}$. We therefore aim to learn a parametric predictor $f_\theta$ by minimizing
\begin{equation}
\mathcal{L}(\theta)
=
\mathbb{E}_{\tau \sim \mathcal{D}}
\left[
\sum_{t=t_0}^{t_0+L-1}
\ell\!\left(
f_\theta(x_{t_0:t}),
\tilde{y}_t
\right)
\right],
\end{equation}
where the predictor outputs the cumulative cytotoxic outcome at time $t$, and $\ell(\cdot,\cdot)$ measures the discrepancy to the ground truth.

\section{Latent NK--Tumor Interaction Dynamics with \ourmodel}
\begin{figure}[t]
    \centering
    \includegraphics[width=1\linewidth]{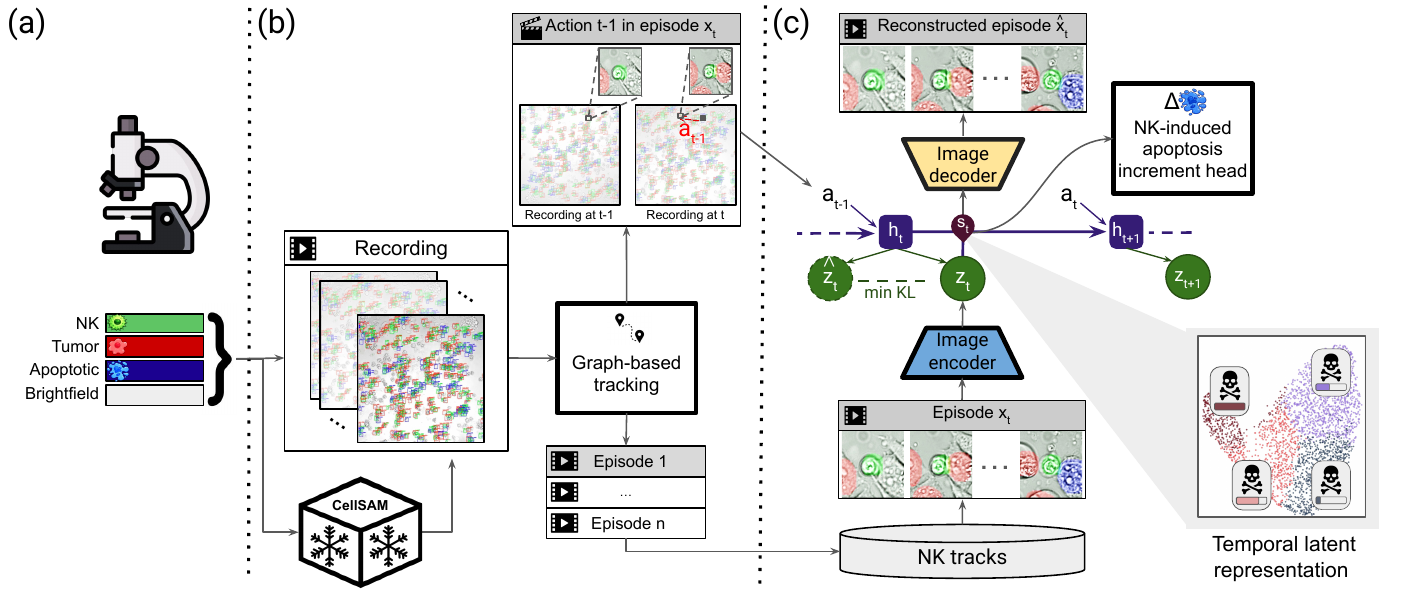}
    \caption{Overview of \textbf{\ourmodel}: (a) Multi-channel fluorescence microscopy captures NK cells, tumor cells, apoptosis, and morphology. (b) Segmentation and tracking yield NK-centered interaction trajectories. (c) \ourmodel encodes these sequences into a recurrent latent state-space model that captures interaction dynamics under partial observability, supports latent rollouts for future cytotoxic outcomes, and predicts NK-induced apoptosis increments that accumulate monotonically. The learned latent space organizes NK behaviors into coherent modes.}
    \label{fig:blink_architecture}
\end{figure}
In this section, we introduce \ourmodel, a latent interaction framework for modeling NK–tumor interaction dynamics and estimating cumulative cytotoxic outcome. Our approach integrates a recurrent state-space world model that captures latent interaction dynamics from time-resolved fluorescence microscopy with a prediction head that estimates per-frame NK-induced apoptosis increments, which are accumulated to produce biologically consistent cumulative cytotoxic outcome trajectories. We describe the latent interaction model, apoptosis increment head, and joint training objective. Fig.~\ref{fig:blink_architecture} provides an overview of the approach.
\subsection{NK Cell World Model Learning}
World models are designed to learn latent dynamics from sequential observations under partial observability. In our setting, the NK cell is treated as the agent interacting with tumor cells, while multi-channel fluorescence microscopy provides partial observations of this biological process. Cytotoxic events arise from latent interaction dynamics that are not directly observable in image space. To model these dynamics, we adopt a recurrent state-space architecture following DreamerV2~\cite{hafnerMasteringAtariDiscrete2020} as the backbone of \ourmodel. The model consists of an image encoder that maps microscopy observations into compact latent features, a recurrent state-space model (RSSM)~\cite{hafnerLearningLatentDynamics2019} for learning interaction dynamics, and a decoder for reconstructing observations from latent states. At each time step, the RSSM maintains a deterministic recurrent state $h_t$ and a stochastic latent state $z_t$, forming the combined model state $s_t = (h_t, z_t)$. Given the previous latent state and the NK cell displacement $a_{t-1}$, the model updates its internal state to obtain the current latent state. The RSSM includes the following components:\looseness=-1
\begin{equation}
\label{modelcomponents}
\begin{aligned}
\text{Recurrent state:}\; 
& h_t = f_\theta(s_{t-1}, a_{t-1}) 
& \text{Representation:}\; 
& z_t \sim q_\theta(z_t \mid h_t, x_t) \\
\text{Dynamics model:}\; 
& \hat{z}_t \sim p_\theta(\hat{z}_t \mid h_t) 
& \text{Decoder:}\; 
& \hat{x}_t \sim p_\theta(\hat{x}_t \mid s_t)
\end{aligned}
\end{equation}
The representation model incorporates the current observation to infer a posterior latent state $z_t$, while the dynamics model learns to approximate this posterior without access to the observation, enabling latent rollouts over extended horizons. The combined model state $s_t$ encodes the evolving latent interaction dynamics. The posterior $q_\theta$ and prior $p_\theta$ are parameterized as categorical distributions and optimized using straight-through gradient estimators~\cite{bengioEstimatingPropagatingGradients2013a}.
\subsection{NK-Induced Apoptosis Increment Head}
To estimate cytotoxic outcome, we attach a prediction head to the latent state $s_t$, implemented as a two-layer MLP. Instead of directly regressing the cumulative tumor cell death, the head predicts a non-negative increment $\lambda_t \geq 0$ via softplus activation, representing expected tumor cell deaths in $(t{-}1, t]$. By construction, we enforce $\lambda_{t_0} = 0$ at each temporal window start. The cumulative prediction within a temporal window starting at $t_0$ is obtained as
\looseness=-1
\begin{equation}
\hat{\tilde{y}}_t = \sum_{\tau = t_0}^{t} \lambda_\tau ,
\end{equation}
which ensures monotonicity by construction. Although supervision is applied to the cumulative signal, the increment-based parameterization enforces non-negativity and induces temporal consistency in cytotoxic outcome.
\subsection{Training Objective}
\ourmodel jointly learns latent interaction dynamics and NK-induced apoptosis increment.
All parameters are optimized end-to-end by minimizing
\begin{equation}
\label{eq:joint_objective}
\mathbb{E}_{\tau \sim \mathcal{D}}
\sum_{t=t_0}^{t_0+L}
\!\left[
-\log p_\theta(x_t \mid s_t)
+ \beta \,\mathrm{KL}\big(
q_\theta(z_t \mid h_t,x_t)
\| p_\theta(\hat{z}_t \mid h_t)
\big)
+ \alpha \,\ell(\hat{\tilde{y}}_t,\tilde{y}_t)
\right]
\end{equation}
where $\beta$ controls KL regularization, $\alpha$ balances latent reconstruction and supervised cytotoxic outcome estimation, and $\ell$ denotes the Huber (smooth L1) loss; we set $\alpha=10$, $\beta=0.3$. Our architecture builds on the DreamerV2~\cite{hafnerMasteringAtariDiscrete2020} latent state-space formulation, following its encoder, decoder, recurrent dynamics, training procedure, and hyperparameters, while adapting supervision and extending it with an apoptosis increment head for cytotoxic outcome estimation.\looseness=-1

\section{Experiments}
We evaluate \ourmodel on time-resolved NK–tumor microscopy sequences to assess its ability to predict cytotoxic outcomes and learn structured behavioral latent representations. Our evaluation has three objectives: (i) determine whether latent dynamical modeling improves cumulative outcome estimation and enables forecasting; (ii) evaluate whether the learned latent space organizes NK trajectories into distinct cytotoxic behavioral modes; and (iii) assess whether inferred behavioral states exhibit coherent temporal transitions consistent with known NK–tumor interaction stages.\looseness=-1
\subsubsection{Dataset:}
We use a long-term time-lapse recording ($\sim$10~h) of NK cells co-cultured with the PC3/PSMA tumor cell line, acquired via synchronized multi-channel fluorescence microscopy. Each frame contains brightfield morphology (Transmission), tumor nuclei (H2B-EGFP), NK cell label (CTFR), and caspase-based viability (NucView405) channels, recorded at 16-bit depth with 60 s temporal resolution, enabling continuous observation of NK--tumor interactions and apoptosis. NK cell trajectories are extracted using CellSAM segmentation~\cite{marksCellSAMFoundationModel2025} and greedy nearest-neighbor tracking based on inter-frame spatial proximity. For each NK track and time step, we generate a $128 \times 128$ NK-centered crop by combining the brightfield image with segmentation masks from the NK, tumor, and viability channels, yielding a pseudo-colored RGB representation of morphology and fluorescence signals. Each frame is paired with a 2D action vector $(\Delta x, \Delta y)$ describing the NK cell’s inter-frame displacement in the imaging plane, and a cumulative cytotoxicity label $c^{(i)}_t$, defined as the cumulative number of NK-induced apoptosis events. Tracks shorter than 60~frames (1~h) are discarded. The remaining trajectories are split into 485 training, 29 validation, and 57 test episodes (85\%/5\%/10\%), with each NK trajectory treated as one episode, yielding approximately 250,000 frames in total. The splits exhibit comparable sequence characteristics: the training set has a mean track length of $430.4 \pm 229.1$ frames and $1.41 \pm 1.19$ outcomes per episode, the validation set has $470.2 \pm 213.0$ frames and $1.55 \pm 1.19$ outcomes, and the test set has $424.6 \pm 231.6$ frames and $1.28 \pm 1.18$ outcomes, indicating a consistent distribution across splits. Across all splits, the number of cytotoxic outcomes per trajectory ranges from 0 to 4.\looseness=-1
\subsubsection{Evaluation Protocol:}
Models are trained on fixed-length windows ($L=50$) sampled from NK trajectories to predict cumulative cytotoxic outcome within each window. At test time, evaluation is performed on full trajectories (up to $L=600$) via sequential rollout. Performance is assessed at the trajectory level using final predicted and ground-truth cumulative outcomes, reporting MAE, RMSE, Pearson correlation, and the percentage of tracks within $\pm 1$ outcome. Future outcome forecasting is evaluated using F-MAE$_{30}$, defined as the mean absolute error over a 30-frame latent rollout without access to future observations. To isolate the contributions of temporal modeling, monotonicity, latent dynamics, and action conditioning, we compare \ourmodel against a hierarchy of baselines trained under identical data splits. We consider: (i) a feedforward autoencoder (FrameAE) without recurrence, assessing whether temporal modeling is necessary; (ii) deterministic recurrent models (GRU-regress and GRU-monotone) without a stochastic latent state or learned prior. GRU-regress directly predicts cumulative outcome, whereas GRU-monotone predicts non-negative increments that are accumulated over time, enforcing monotonicity by construction. Lacking a learned latent prior, these models cannot perform reliable latent forecasting; and (iii) an observation-only recurrent state-space model without action input, which retains stochastic latent dynamics and the same monotonic increment head, isolating the contribution of action conditioning. All models share the same encoder architecture, optimizer, and training protocol to ensure a fair comparison.\looseness=-1
\begin{table}[t]
\caption{Track-level cumulative cytotoxic outcome prediction on the held-out test set, showing improvements of \ourmodel\ across error and forecasting metrics.}
\centering
\small
\setlength{\tabcolsep}{2.pt}
\begin{tabular}{lccccc}
\toprule
Model 
& MAE $\downarrow$ 
& RMSE $\downarrow$ 
& Corr $\uparrow$ 
& Within $\pm1$ (\%) $\uparrow$
& F-MAE$_{30}$ $\downarrow$ \\
\midrule
Zero
& 1.28$\pm$0.16 & 1.74$\pm$0.15 & 0$\pm$0.0 & 54.3\%$\pm$7.0\% & 0.12$\pm$0.06 \\

Mean 
& 1.04$\pm$0.07 & 1.18$\pm$0.08 & 0$\pm$0.0 & 49.6\%$\pm$6.3\% & 0.24$\pm$0.05\\

FrameAE
& 0.95$\pm$0.11 & 1.14$\pm$0.13 & 0.32$\pm$0.07 & 64.9\%$\pm$6.8\% & X \\

GRU-regress 
& 1.25$\pm$0.14 & 1.72$\pm$0.14 & 0$\pm$0.0 & 55.7\%$\pm$6.9\% & 0.12$\pm$0.06 \\

GRU-monotone 
& 0.74$\pm$0.09 & 1.04$\pm$0.11 & 0.57$\pm$0.04 & 71.9\%$\pm$3.3\% & 0.22$\pm$0.04 \\

\ourmodel-no-action 
& 0.80$\pm$0.06 & 1.14$\pm$0.09 & 0.61$\pm$0.04 & 69.4\%$\pm$7.3\% & 0.09$\pm$0.01 \\

\textbf{\ourmodel} 
& \textbf{0.60$\pm$0.07} & \textbf{0.81$\pm$0.08} & \textbf{0.77$\pm$0.05} & \textbf{80.7\%$\pm$5.2\%} & \textbf{0.05$\pm$0.01} \\
\bottomrule
\end{tabular}
\label{tab:main_results}
\end{table}

\begin{figure}[t]
    \centering
    \includegraphics[width=1\linewidth]{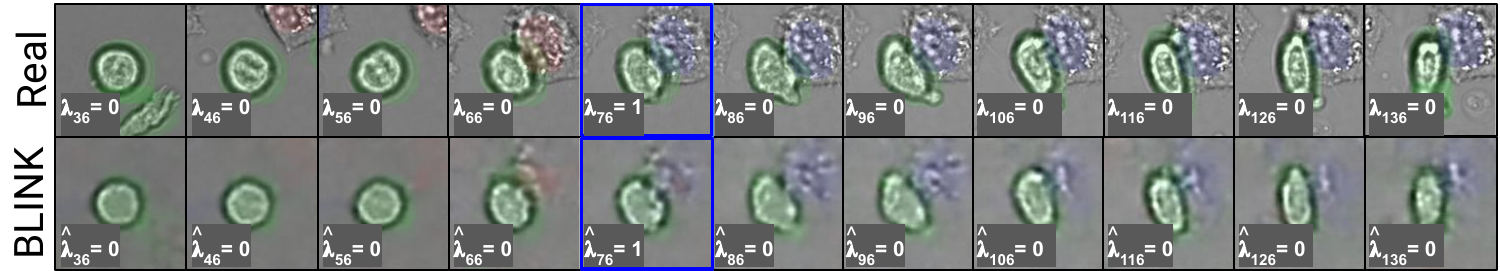}
    \caption{Real interaction trajectory (top) and world model-decoded latent trajectory (bottom). Predicted and ground truth apoptosis increments align.}
    \label{fig:real_wm}
\end{figure}

Table \ref{tab:main_results} reports track-level performance on the held-out test set, where \ourmodel\ consistently outperforms all baselines. While FrameAE improves over Zero and Mean, the strong gain of GRU-monotone over FrameAE highlights the importance of temporal modeling. In contrast, GRU-regress collapses to the trivial zero predictor due to sparse cytotoxic events, underscoring the need for monotonic constraints. Comparing GRU-monotone with BLINK-no-action, we observe comparable outcome accuracy, with BLINK showing slightly higher MAE but substantially stronger forecasting. This trade-off is expected: the stochastic recurrent state-space model is trained to jointly reconstruct observations and regularize latent dynamics, thereby learning a prior over interaction evolution. While this broader objective does not exclusively optimize supervised outcome error, it enables coherent future rollouts and structured latent transitions. In contrast, deterministic baselines lack a learned latent transition prior and cannot perform true latent forecasting; GRU predictions rely on deterministic hidden-state propagation, and FrameAE cannot be rolled out beyond observed inputs. Finally, when augmenting \ourmodel with action conditioning, performance improves across both final outcome prediction and Forecast-MAE$_{30}$, demonstrating that structured latent dynamics combined with explicit modeling of NK motion yields the most accurate and temporally consistent characterization of cytotoxic behavior. As shown in Fig.~\ref{fig:real_wm}, the latent world model captures interaction dynamics and produces increment predictions consistent with observed cytotoxic events.\looseness=-1
\begin{figure}[t]
\centering
\begin{subfigure}[t]{0.30\columnwidth}
    \centering
    \includegraphics[width=\linewidth]{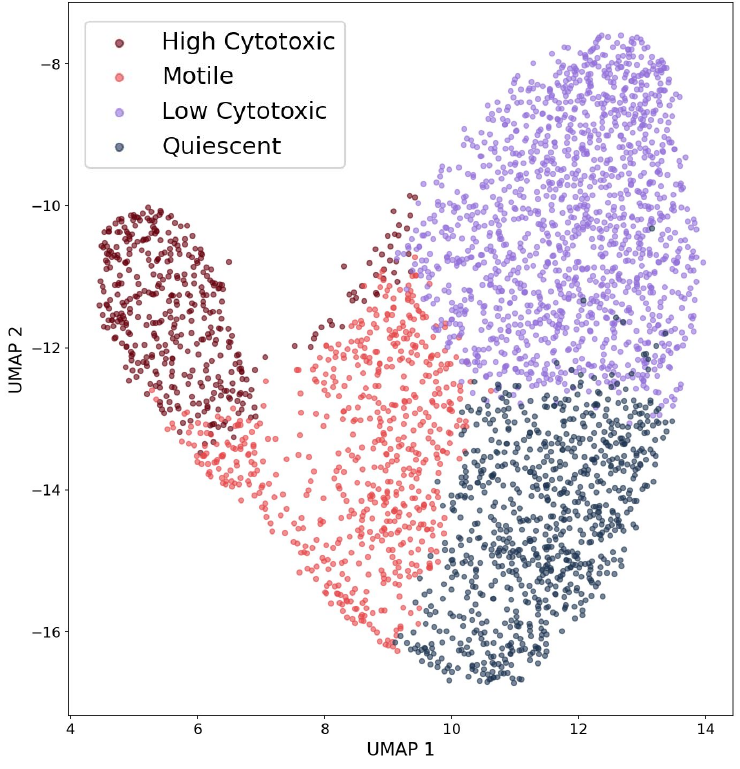}
    \caption{}
    \label{fig:umap_a}
\end{subfigure}
\hfill
\begin{subfigure}[t]{0.30\columnwidth}
    \centering
    \includegraphics[width=\linewidth]{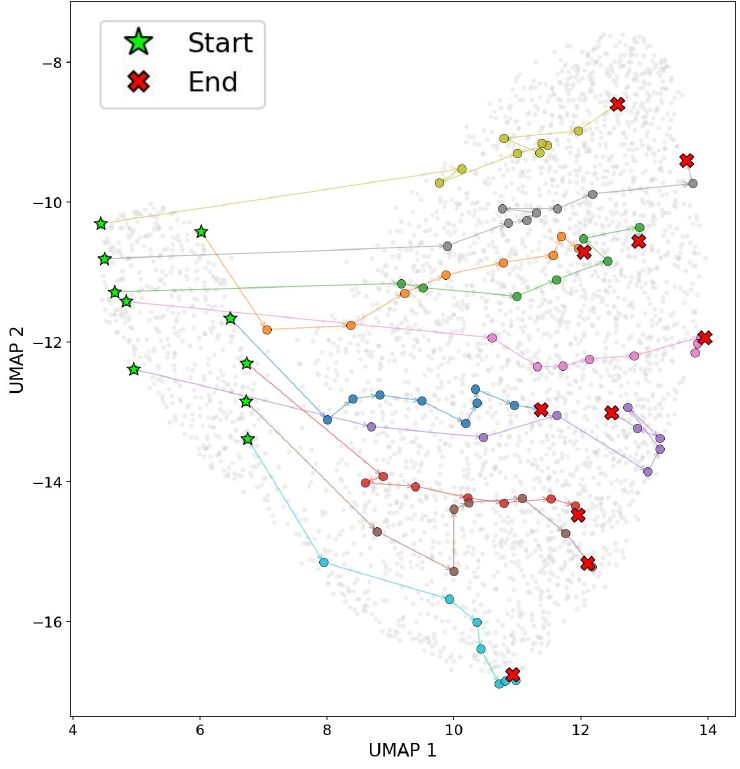}
    \caption{}
    \label{fig:umap_b}
\end{subfigure}
\hfill
\begin{subfigure}[t]{0.38\columnwidth}
    \centering
    \includegraphics[width=\linewidth]{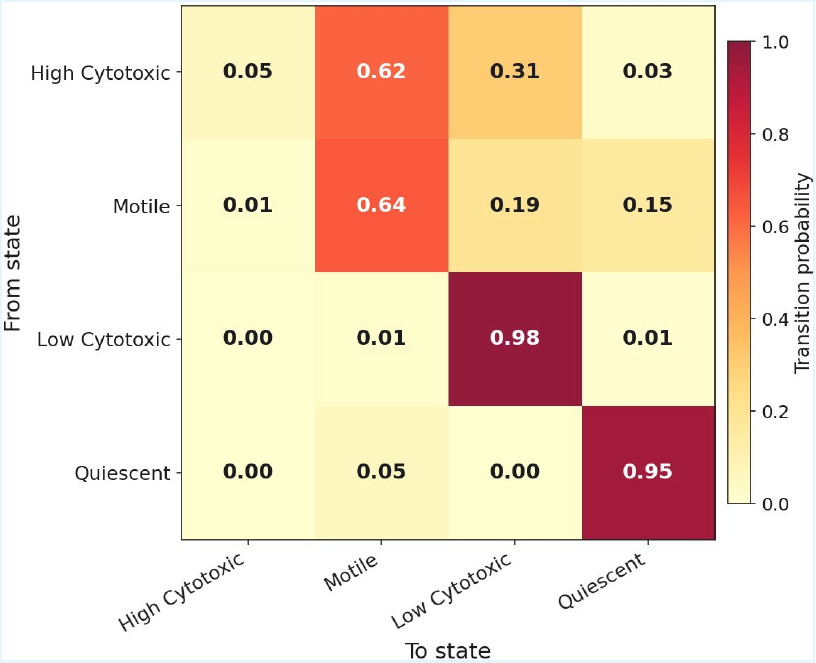}
    \caption{}
    \label{fig:umap_c}
\end{subfigure}
\caption{
Latent behavioral structure of NK trajectories.
(a) UMAP of training window embeddings clustered into four modes.
(b) Test tracks projected into the embedding.
(c) State transition matrix showing temporal mode progression.
}
\label{fig:behavior_umap}
\end{figure}
To evaluate whether the learned latent space organizes NK trajectories into distinct cytotoxic behavioral modes and coherent temporal progression (Fig.~\ref{fig:behavior_umap}), we extracted per-frame latent states from training tracks and constructed sliding-window embeddings (length=30, stride=30) by aggregating the mean and temporal change of latent features within each window. The embeddings were standardized, reduced with PCA, and clustered unsupervised into four groups using KMeans. Characterization by window-level cytotoxic outcome and migration speed revealed four separable states: High Cytotoxic (mean outcome: 0.56, mean speed: 5.60; 12.9\% of windows), Motile (0.26, 5.67; 19.2\%), Low Cytotoxic (0.13, 1.55; 43.0\%), and Quiescent (0.09, 1.44; 24.9\%). The clear differences in outcome and motility across clusters indicate that the latent space captures functionally distinct cytotoxic regimes rather than arbitrary partitions (Fig.~\ref{fig:umap_a}). Held-out test tracks projected into the embedding (Fig.~\ref{fig:umap_b}) follow structured paths across these regions, starting in High Cytotoxic and ending in Low Cytotoxic or Quiescent states. The transition matrix on the test set (Fig.~\ref{fig:umap_c}) shows preferential flows from High Cytotoxic to Motile and subsequently to Low Cytotoxic or Quiescent states, consistent with progressive engagement, cytotoxic outcome, and decline phases of NK–tumor interactions. Overall, \ourmodel improves cumulative outcome prediction, enables forecasting, and learns an interpretable latent representation with structured behavioral mode progression.\looseness=-1

\section{Conclusion}
We presented \ourmodel, a trajectory-based latent world model for estimating cumulative NK cytotoxic outcome from time-resolved fluorescence microscopy. By formulating cytotoxicity as inference over partially observable interaction states, \ourmodel enables grounded prediction beyond frame-wise classification. Our action-conditioned recurrent state-space model with monotonic increments supports forecasting and, on long-term NK–tumor recordings, uncovers coherent behavioral modes. Together, these results demonstrate that NK cytotoxic outcome can be modeled as a latent dynamical process at single-cell resolution.\looseness=-1
\begin{credits}
\subsubsection{\ackname} The authors gratefully acknowledge financial support from the German Research Foundation (DFG, Deutsche Forschungsgemeinschaft) – Project-ID 499552394 – CRC 1597 “SmallData”, as well as Project-ID UL316/9-1 (to E.U. and A.M.) and SFB/IRTG 1292 (Project-ID 318346496 to E.U. and A.M.). Additional support was provided by the German Cancer Aid (Stiftung Deutsche Krebshilfe) within the framework of preCDD/CAR Factory (ID: 70115200) and by the Mertelsmann Foundation. This work was also partly funded as part of BrainLinks-BrainTools, which is supported by the Federal Ministry of Economics, Science and Arts of Baden-Württemberg within the sustainability program for projects of the Excellence Initiative II.

\subsubsection{\discintname}
Evelyn Ullrich has a sponsored research project with Gilead and BMS and acts as medical advisor of Phialogics and CRIION.
\end{credits}
\bibliographystyle{splncs04}
\bibliography{sections/references}

\end{document}